\documentclass[11pt]{article}
\usepackage{coling2018}
\usepackage{times}
\usepackage{url}
\usepackage{latexsym}

\usepackage{graphicx}
\usepackage{tabularx}
\usepackage{soul}
\usepackage{epstopdf}
\usepackage{longtable}
\usepackage{tabu}
\usepackage{xargs}
\usepackage{lipsum}
\usepackage[latin1]{inputenc}
\usepackage{xstring}
\usepackage{todonotes}
\usepackage{booktabs}
\usepackage{enumitem}
\usepackage{amsmath}
\usepackage{makecell}
\usepackage{url}
\usepackage{breakurl}

\title{Folksonomication: Predicting Tags for Movies from Plot Synopses Using Emotion Flow Encoded Neural Network}

\author{Sudipta Kar \qquad Suraj Maharjan \qquad Thamar Solorio \\
Department of Computer Science\\
University of Houston\\
Houston, TX 77204-3010\\
{\tt { \{skar3, smaharjan2, tsolorio\}}@uh.edu}}

\date{}

\begin{document}
\maketitle
\begin{abstract}
Folksonomy of movies covers a wide range of heterogeneous information about movies, like the genre, plot structure, visual experiences, soundtracks, metadata, and emotional experiences from watching a movie. 
Being able to automatically generate or predict tags for movies can help recommendation engines improve retrieval of similar movies, and help viewers know what to expect from a movie in advance. In this work, we explore the problem of creating tags for movies from plot synopses. We propose a novel neural network model that merges information from synopses and emotion flows throughout the plots to predict a set of tags for movies. We compare our system with multiple baselines and found that the addition of emotion flows boosts the performance of the network by learning $\approx$18\% more tags than a traditional machine learning system.
\end{abstract}

%
%
\blfootnote{
    %
    %
    \hspace{-0.65cm}  
    %
    %
    
    
    \hspace{-0.65cm}  
    This work is licensed under a Creative Commons 
    Attribution 4.0 International License.
    License details:
    \url{http://creativecommons.org/licenses/by/4.0/}
}
\section{Introduction}\label{sec:intro}
User generated tags for online items are beneficial for both of the users and content providers in modern web technologies.
For instance, the capability of tags in providing a quick glimpse of items can assist users to pick items precisely based on their taste and mood.
On the other hand, such strength of tags enables them to act as strong search keywords and efficient features for recommendation engines \cite{Lambiotte2006,szomszor2007folksonomies,li2008tag,borne2013collaborative}. 
As a result, websites for different medias like photography\footnote{\url{http://www.flickr.com}}, literature\footnote{\url{http://www.goodreads.com}}, film\footnote{\url{http://www.imdb.com}}, and music\footnote{\url{http://www.last.fm}} have adopted this system to make information retrieval easier. 
Such systems are often referred as Folksonomy \cite{vander2005folksonomy}, social tagging, or collaborative tagging.\\
%
%
%
%
In movie review websites, it is very common that people assign tags to movies after watching them.
Tags for movies often represent summarized characteristics of the movies such as emotional experiences, events, genre, character types, and psychological impacts.
As a consequence, tags for movies became remarkably convenient for recommending movies to potential viewers based on their personal preferences and user profiles. 
However, this situation is not the same for all of the movies.
Popular movies usually have a lot of tags as they tend to reach a higher number of users in these sites.
On the other hand, low profile movies that fail to reach such an audience have very small or empty tagsets. 
In an investigation, we found that $\approx$34\% of the movies among the top $\approx$130K movies of 22 genres\footnote{\url{http://www.imdb.com/genre/}} in IMDB do not have any tag at all.
It is very likely that lack of descriptive tags negatively affects chances of movies being discovered.\\
An automatic process to create tags for movies by analyzing the written plot synopses or scripts could help solve this problem. Such a process would reduce the dependency on humans to accumulate tags for movies. Additionally, learning the characteristics of a movie plot and possible emotional experiences from the written synopsis is also an interesting problem by itself from the perspective of computational linguistics. 
%
%
As the attributes of movies are multi-dimensional, a tag prediction system for movies has to generate multiple tags for a movie. 
The application of predicting multiple tags from textual description is not necessarily limited to the domain of movie recommendation but also
appropriate in other domains, such as video games and books, where storytelling is relevant.
In this paper, we explore the problem of analyzing plot synopses to generate multiple plot-related tags for movies. Our key contributions in this paper are as follows:
\begin{itemize}[leftmargin=0.2cm,labelindent=0cm,noitemsep]
\item We create a neural system for predicting tags from narrative texts and provide a robust comparison against traditional machine learning systems. Table \ref{tab_example_tag_fp} shows examples of predicted tags by our system for four movies.
\item We propose a neural network model that encodes flow of emotions in movie plot synopses. This emotion flow helps the model to learn more attributes of movie plots.
\item We release our source code and a live demo of the tag prediction system at\\ \url{http://ritual.uh.edu/folksonomication-2018}.
\end{itemize}
\begin{table}[!hbtp]
\small
\centering
\begin{tabular}{|c|c|c|}
\hline
\textbf{IMDB ID} & \makecell{\textbf{Movie Title}} & \makecell{\textbf{Predicted Tags}}\\ \hline

tt0133093 &
\makecell{The Matrix} &
\makecell{\textcolor{blue}{though-provoking}, \textcolor{blue}{action}, \textcolor{blue}{sci-fi}, \textcolor{red}{suspenseful}, \textcolor{blue}{mystery}} \\ 
\hline

tt0233298 &
\makecell{Batman Beyond: Return of the Joker} &
\makecell{\textcolor{red}{action}, \textcolor{blue}{good versus evil}, \textcolor{blue}{suspenseful}, \textcolor{red}{humor}, \textcolor{red}{thought-provoking}}\\
\hline

tt0309820 &
\makecell{Luther} &
\makecell{\textcolor{blue}{murder}, \textcolor{red}{melodrama}, \textcolor{red}{intrigue}, \textcolor{red}{historical fiction}, \textcolor{red}{christian film}} \\
\hline

tt0163651 &
\makecell{American Pie} &
\makecell{\textcolor{blue}{adult comedy}, \textcolor{red}{cute}, \textcolor{red}{feel-good}, \textcolor{red}{prank}, \textcolor{blue}{entertaining}} \\
\hline

\end{tabular}
\label{tab_example_tag_fp}
\caption{\small Example of predicted tags from the plot synopses of four movies. Blue and red labels indicate true positives and false positives respectively.}\label{tab:tab_tags_script_plot_example}
\end{table}

\section{Related Work}\label{sec:related_works}
Automatic tag generation from content-based analysis has drawn attention in different  domains like music and images.
For example, creating tags for music has been approached by  utilizing lyrics~\cite{music_mood_lyrics_zaanen_2010,music_lyrics_mining_hu_2009}, acoustic features from the tracks~\cite{music_social_tag_nips_2008,music_mscale_dieleman}, categorical emotion models~\cite{music_mood_cls_kim2001}, and deep neural models~\cite{music_cnn_choi_2017}.\\ 
%
AutoTag \cite{autotag_text_Mishne_2006} and TagAssist \cite{tagassist_sood_2007}, which utilize the text content to generate tags, aggregate information from similar blog posts to compile a list of ranked tags to present to the authors of new blog posts.
Similar works \cite{KTV08,bib_tag_Lipczak08tagrecommendation,bib_tag_tatu2008challenge} focused on recommending tags to users of BibSonomy\footnote{\url{https://www.bibsonomy.org}} upon posting a new web page or publication as proposed systems in the ECML PKDD Discovery Challenge 2008 \cite{rsdc_2008} shared task. 
These systems made use of some kind of out of content resources like user metadata, and tags assigned to similar resources to generate tags.\\
Computational narrative studies deal with representing natural language stories by computational models that can be useful to understand, represent, and generate stories computationally. Current works attempt to model narratives using the character's personas and roles \cite{char-role-iden-folk-valls,bamman2014learning}, interaction information between the characters \cite{feudin_family:iyyer_naacl,ChaturvediSDD16,chaturvedi2017unsupervised} and  events taking place throughout the stories \cite{plot_unit_goyal,folktale_finlayson,plot_induction_mcintyre-lapata:2010:ACL}. Other works try to build social networks of the characters \cite{agarwal-socialnet1,agarwal-socialnet2,agarwal-socialnet3,socialnet-krishnan}. 
Only a few works explored the possible type of impressions narrative texts can create on their consumers.
For instance, different types of linguistic features have been used for success prediction for books \cite{success_novel_ganjigunteashok_feng_choi:2013:EMNLP,book_likeability_maharjan2017multi} and tag prediction of movies from plot synopses \cite{mpst_2018}.
The tag prediction system predicts a fixed number of tags for each movie.
But the tag space created by the system for the test data covers only 73\% tags of the actual tagset as the system could capture a small portion of the multi-dimensional attributes of movie plots.

\section{Dataset}\label{sec:dataset}
We conduct our experiments on the Movie Plot Synopses with Tags (MPST) corpus \cite{mpst_2018}, which is a collection of plot synopses for 14,828 movies collected from IMDb and Wikipedia.
Most importantly, the corpus provides one or more fine-grained tags for each movie.
The reason behind selecting this particular dataset is two-fold.
First, the tagset is comprised of manually curated tags. 
These tags express only plot-related attributes of movies (e.g. suspenseful, violence, and melodrama) and are free of any tags foreign to the plots, such as metadata. 
Furthermore, grouping semantically similar tags and representing them by generalized tags helped to reduce the noise created by redundancy in tag space.
Second, the corpus provides adequate amount of texts in the plot synopses as all the synopses have at least ten sentences.
We follow the same split provided with the corpus, using 80\% for training and 20\% for test set. Table \ref{tab:mpst_stat} gives statistics of the dataset.
\begin{table}[!htbp]
\centering
\begin{tabular}{l|ccccc}
\toprule
\hline
Split & \#Plot Synopses & \#Tags & \#Tags per Movie & \#Sentence per Synopsis & \#Words per Synopsis \\ \hline
Train          & 11862                    & 71              & 2.97                      & 42.36                            & 893.39                        \\
Test           & 2966                     & 71              & 3.04                      & 42.61                            & 907.96                        \\ \hline
\end{tabular}
\caption{Statistics of the MPST corpus.}
\label{tab:mpst_stat}
\end{table}

\section{Encoding Emotion Flow with a Neural Network}\label{sec:methodology}
\begin{figure}[t]
\centering
\includegraphics[scale=0.4]{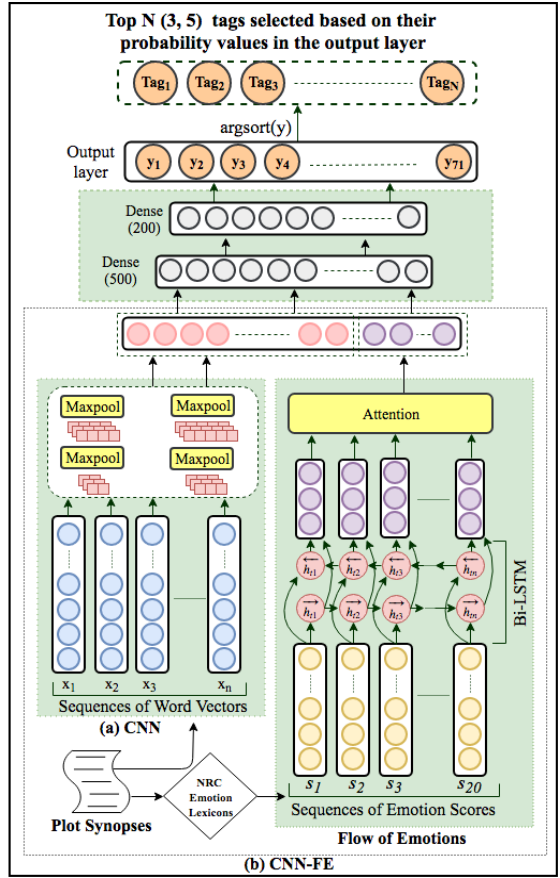}
\caption{\small Convolutional Neural Network with Emotion Flow. The entire model is a combination of three modules. Module (a) learns feature representations from synopses using convolutional neural network. Module (b) incorporates emotion flows with module (a) to generate a combined representation of synopses. Module (c) uses these representations to predict the likelihood of each tag.}
\label{fig_main_model}
\end{figure}

\setlength{\abovedisplayskip}{1pt}
\setlength{\abovedisplayshortskip}{1pt}
\setlength{\belowdisplayskip}{1pt}
\setlength{\belowdisplayshortskip}{1pt}
Our proposed model simultaneously takes the emotion flow throughout the storyline and the text-based representation of the synopsis to retrieve relevant tags for a movie. 
Figure \ref{fig_main_model} shows the proposed architecture. 
The proposed neural architecture has three modules. 
The first module uses a convolutional neural network (CNN) to learn plot representations from synopses. 
The second module models the flow of emotions via a bidirectional long short-term memory (Bi-LSTM) network. 
And the last module contains hidden dense layers that operate on the combined representations generated by the first and second modules to predict the most likely tags for movies.
%
~\\
\noindent{\textbf{(a) Convolutional Neural Network (CNN)}}:
Recent successes in different text classification problems motivated us to extract important word level features using  convolutional neural networks (CNNs) \cite{conv1_santos,Kim2014,zhang2015character,Kar2017,Shrestha2017}. 
We design a model that takes word sequences as input, where each word is represented by a 300-dimensional word embedding vector.
We use randomly initialized word embeddings but also experiment with the FastText\footnote{\url{https://fasttext.cc/docs/en/english-vectors.html}} word embeddings trained on Wikipedia using subword information. 
We stack 4 sets of one-dimensional convolution modules with 1024 filters each for filter sizes 2, 3, 4, and 5 to extract word-level $n$-gram features~\cite{Kim2014,zhang2015character}. 
Each filter of size \textit{c} is applied from window \(t\) to window $t+c-1$ on a word sequence \(x_1, x_2, \ldots ,x_n\). 
Convolution units of filter size \(c\) calculate a convolution output using a weight map \(W_c\), bias \(b_c\), and the ReLU activation function \cite{Nair:2010:RLU:3104322.3104425}. The output of this operation is defined by:
\begin{align}
	h_{c,t} = ReLU(W_c x_{t:t+c-1} + b_c)
\end{align}
The ReLU activation function is defined by:
\begin{align}
	ReLU(x) = max(0, x)
\end{align}
Finally, each convolution unit produces a high-level feature map  $h_c$.
\begin{equation}
	h_c = [h_{c,1}, h_{c,2}, ..., h_{c,T-c+1},]
\end{equation}
On those feature maps, we apply max-over-time pooling operation and take the maximum value as the feature produced a particular filter. 
We concatenate the outputs of the pooling operation for four filter sets that represent the feature representations for each plot synopsis. \\
~\\
\noindent{\textbf{(b) CNN with Flow of Emotions (CNN-FE): }}
\textcolor{black}{
Stories can be described in terms of emotional shapes \cite{sunday1981autobiographical}, and it has been shown that the emotional arcs of stories are dominated by six different shapes \cite{emotional_arc_reagan2016}. We believe that capturing the emotional ups and downs throughout the plots can help  better understand how the story unfolds. This will enable us to predict relevant tags more accurately. So we design a neural network architecture that tries to learn representations of plots using the vector space model of words combined with the emotional ups and downs of plots.}\\
Human emotion is a complex phenomenon to define computationally.
The Hourglass of Emotions model \cite{hourglass_emotion} categorized human emotions into four affective dimensions (\textit{attention, sensitivity, aptitude, and pleasantness}), which started from the study of human emotions by \newcite{plutchik2001nature}. 
Each of these affective dimensions is represented by six different activation levels that make up to 24 distinct labels called `elementary emotions' that represent the total emotional state of the human mind. 
NRC\footnote{National Research Council Canada} emotion lexicons \cite{emolex1} is a list of 14,182 words\footnote{Version 0.92} and their binary associations with eight types of elementary emotions from the Hourglass of Emotions model (\textit{anger, anticipation, joy, trust, disgust, sadness, surprise,} and \textit{fear}) with polarity. 
These lexicons have been used effectively in tracking the emotions in literary texts \cite{emotrack} and predicting success of books \cite{maharjan-EtAl:2018:N18-2}.\\
To model the flow of emotions throughout the plots, we divide each synopsis into $N$ equally-sized segments based on words.
For each segment, we compute the percentage of words corresponding to each emotion and polarity type (positive and negative) using the NRC emotion lexicons. 
More precisely, for a synopsis $x \epsilon X$, where $X$ denotes the entire collection of plot synopses, we create $N$ sequences of emotion vectors using the NRC emotion lexicons as shown below:
\begin{equation}\label{eq:seq}
	x \rightarrow s_{1:N} = [s_{1}, s_{2}, ..., s_{N}]
\end{equation}
where $s_{i}$ is the emotion vector for segment $i$.
We experiment with different values of $N$, and $N=20$ works better on the validation data.\\
As recurrent neural networks are good at encoding sequential data, we feed the sequence of  emotion vectors into a bidirectional LSTM~\cite{hochreiter:1997} with 16 units as shown in Figure \ref{fig_main_model}. This bidirectional LSTM layer tries to summarize the contextual flow of emotions from both directions of the plots. The forward LSTMs read the sequence from $s_{1}$ to $s_{N}$, while the backward LSTMs read the sequence in reverse from $s_{N}$ to $s_{1}$. These operations will compute the forward hidden states  $(\overrightarrow{h_{1}}, \ldots, \overrightarrow{h_{N}})$ and backward hidden states $(\overleftarrow{h_{1}}, \ldots, \overleftarrow{h_{N}})$. 
For input sequence $s$, the hidden states $h_{t}$ are computed using the following intermediate calculations:
\vspace*{-0.1mm}
\begin{align*}
\hspace*{4mm}
i_{t} &= \sigma(W_{si}s_{t} + W_{hi}h_{t-1} + W_{ci}c_{t-1} + b_{i})\\
f_{t} &= \sigma(W_{sf}s_{t} + W_{hf}h_{t-1} + W_{cf}c_{t-1} + b_{f})\\
c_{t} &= f_{t}c_{t-1} + i_{t} \tanh(W_{sc}s_{t} + W_{hc}h_{t-1} + b_{c})\\
o_{t} &= \sigma (W_{sc}s_{t} + W_{hc}h_{t-1} + b_{c})\\
h_{t} &= o_{t} \tanh(c_t)
\end{align*}
where, $W$ and $b$ denote the weight matrices and bias, respectively. $\sigma$ is the sigmoid activation function, and $i$, $f$, $o$, and $c$ are \textit{input gate, forget gate, output gate}, and \textit{cell} activation vectors, respectively. 
The annotation for  each segment $s_{i}$ is obtained by concatenating its forward hidden states $\overrightarrow{h_{i}}$ and backward hidden states $\overleftarrow{h_{i}}$, i.e. $h_{i}$=$[\overrightarrow{h_{i}}; \overleftarrow{h_{i}}]$. 
We then apply attention mechanism on this representation to get a unified representation of the emotion flow.\\
Attention models have been used effectively in many problems related to computer vision \cite{attention:MnihHGK14,attention:BaMK14}  and have been  successfully adopted in problems  related to natural language processing \cite{DBLP:journals/corr/BahdanauCB14,attention:SeoLCSH16}. An attention layer applied on top of a feature map $h_i$ computes the weighted sum $r$ as follows:
\begin{equation}\label{eq:attention}
	r = \sum_{i}\alpha _{i} h_{i}
\end{equation}
and the weight $\alpha_{i}$ is defined as
\begin{equation}
    \alpha_{i}=\frac{\exp({score(h_{i})})}{\sum_{{i}'}\exp(score(h_{{i}'}))},
\end{equation}
where, $score(.)$ is computed as follows:
\begin{equation}
score(h_{i})= v^{T}tanh(W_{a}h_{i}+b_a)
\end{equation}
where, $W$, $b$, $v$, and $u$ are model parameters.
Finally, we concatenate the representation of the emotion flow produced by the attention operation and the output vector with the vector representation generated from the CNN module.\\
%
%
The concatenated vector is then fed into two hidden dense layers with 500 and 200 neurons. To improve generalization of the model, we use dropout with a rate of 0.4 after each hidden layer. Finally, we add the output layer $\hat{y}$ with 71 neurons to compute predictions for 71 tags. To overcome the imbalance of the tags, we weight the posterior probabilities for each tag using different weight values. Weight value $CW_t$ for tag $t \epsilon T$ is defined by,
\begin{equation}
	CW_t = \frac{ |D|}{|T| \times M_t}
\end{equation}
where, $|D|$ is the size of the training set, $|T|$ is the number of classes, and $M_t$ is the number of movies having tag $t$ in the training set.
%
We normalize the output layer by applying a softmax function defined by,
\begin{equation}
	softmax(\hat{y}) = \frac{ exp(\hat{y}) }{ \sum_{k=0}^{70} exp(\hat{y_{k}})}
\end{equation}
Based on the ranking for each tag, we then select top $N$ (3/5/10) tags  for a movie.\\
%

\section{Experimental Setup}\label{sec:exp_setup}
\noindent{\textbf{Data Processing and Training: }}As a preprocessing step, we lowercase the synopses, remove stop-words and also limit the vocabulary to top 5K words to reduce noise and data sparsity. Then we convert each synopsis into a sequence of 1500 integers where each integer represents the index of the corresponding word in the vocabulary. For the sequences longer than 1500 words, we truncate them from the left based on experiments on the development set. Shorter sequences are left padded with zeros.\\
During training, we use 20\% of the training data as validation data. 
We tune various deep model parameters (dropouts, learning rate, weight initialization schemes, and batch size) using  early stopping technique on the validation data. 
We use the Kullback-Leibler (KL) divergence \cite{kullback1951} to compute the loss between the true and predicted tag distributions and train the network using the \mbox{RMSprop} optimization algorithm \cite{rmsprop_tieleman2012lecture} with a learning rate of 0.0001. We implemented our neural network using the PyTorch deep learning framework\footnote{\url{https://pytorch.org}}.\\
\noindent \textbf{Baselines:} We compare the model performance against three baselines:  majority baseline, random baseline, and traditional machine learning system.
The majority baseline method assigns the most frequent three or  five or ten tags in the training set to all the movies.
Similarly, the random baseline assigns randomly selected three or five or ten tags to each movie. 
Finally, we compare our results with the benchmark system reported in \newcite{mpst_2018}.
This benchmark system used different types of hand-crafted lexical, semantic, and sentiment features to train a OneVsRest approach model with logistic regression as the base classifier.\\
\noindent \textbf{Evaluation Measures:}
We try to follow the same evaluation methodology as described in \newcite{mpst_2018}. 
We create two sets of tags for each movie by choosing the most likely three and five tags by the system.
Additionally, we report our results on a wider range of tags, where we select top ten predictions.
We evaluate the performance using the number of unique tags learned by the system (TL), micro averaged F1, and tag recall (TL).
Tags learned (TL) computes how many unique tags are being predicted by the system for the test data (size of the tag space created by the model for test data).
Tag recall represents the average recall per tag and it is defined by the following equation:
\begin{equation}
TR = \frac{\sum_{i=1}^{|T|}|R_i|}{|T|}
\label{eq:tr}
\end{equation}
Here, $|T|$ is the total number of tags in the corpus, and $R_i$ is the recall for the $i^{th}$ tag.
\\
%
%

\section{Results and Discussions}\label{sec:analysis}
\begin{table}[h]
\centering
\small
\begin{tabular}{@{}|l|rrr|rrr|rrr|@{}}
\hline
 & \multicolumn{3}{c|}{\textbf{Top 3}} & \multicolumn{3}{c|}{\textbf{Top 5}} & \multicolumn{3}{c|}{\textbf{Top 10}} \\  \hline
 
 \textbf{Methods} & \textbf{TL} & \textbf{F1} & \textbf{TR}& \textbf{TL}  & \textbf{F1} & \textbf{TR} & \textbf{TL}  & \textbf{F1} & \textbf{TR}  \\  \hline
 
Baseline: Most Frequent & 3 & 29.7 & 4.23 & 5 & 28.4 & 14.08 & 10 & 28.4 & 13.73 \\
Baseline: Random & 71 & 4.2 & 4.21 & 71 & 6.4 & 15.04 & 71 & 6.6 & 14.36\\ \hline
Baseline: \newcite{mpst_2018} & 47 & \textbf{37.3} & \textbf{10.52} & 52 & \textbf{37.3} & \textbf{16.77} & --- & --- & --- \\  \hline \hline

CNN without class weights &  24 & 36.8 & 7.99 & 26 &36.7 & 12.62 & 27 & \textbf{31.3} & 24.52\\ 
CNN with class weights & 49 & 34.9 & 9.85 & 55 & 35.7 & 14.94 & 67 & 30.8 & \textbf{26.86}  \\ 

CNN-FE & \textbf{58} & 36.9 & 9.40 & \textbf{65} & 36.7 & 14.11 & \textbf{70} & 31.1& 24.76\\ 

\makecell[l]{CNN-FE + FastText} & 53 & \textbf{37.3} & 10.00 &59  & 36.8 & 15.47 &63  & 30.6 & 26.45  \\  \hline

\end{tabular}%
\caption{\small Performance of tag prediction systems on the test data. We report results of two setups using three matrices (TL: Tags learned, F1: Micro f1, TR: Tag recall).}\label{tab:tab_results_test}
\end{table}
Table \ref{tab:tab_results_test} shows our results for Top 3, Top 5, and Top 10 settings.
We will mainly discuss the results achieved by selecting top five tags as it allows us to compare with all the baseline systems and more tags to discuss about.
As the most frequent baseline system assigns a fixed set of tags to all the movies, it fails to exhibit diversity in the created tag space.
Still it manages to achieve a micro-F1 score around 28\%.
On the other hand, the random baseline system creates the most diverse tag space by using all of the possible tags.
However its lower micro-F1 score of 6.30\%  makes it impractical to be used in real world scenario. 
At this point, we find an interesting trade-off between accuracy and diversity.
It is expected that a good movie tagger will be able to capture the multi-dimensional attributes of the plots that allows to generalize a diverse tag space.
Tagging a large collection of movies with a very small and fixed set of tags (e.g. majority baseline system) is not  useful for either a recommendation system or users.
Equally important is the relevance between the movies and the tags created for those movies.
The hand-crafted features based approach \cite{mpst_2018} achieves a micro-F1 around 37\%, which outperforms the majority and random baselines. 
But the system was able to learn only 52 tags, which makes 73\% of the total tags.\\
Our approach achieves a lower micro-F1 score than the traditional machine learning one, but it performs better in terms of learning more tags.
We observe that the micro-F1 of the CNN model with only word sequences is very close (36.7\%) to the hand-crafted features based system. 
However, it is able to learn only around 37\% of the tags.
By utilizing class weights in this model (see Eq. 8), we improve the learning for under-represented tags yielding an increase in \textit{tag recall} (TR) and \textit{tags learned} (TL). 
But the micro-f1 drops  to 35.7\%. 
With the addition of emotion flows to  CNN, the CNN-FE model learns significantly more tags while micro-F1 and tag recall do not change much.
Initializing the embedding layer with pre-trained embeddings made a small improvement in micro-F1 but the model learns comparatively lesser tags.
If we compare the CNN-FE model with the hand-crafted feature based system, micro-F1 using CNN-FE is slightly lower ($\approx 1\%$) than the feature based system. But it provides a strong improvement in terms of the number of tags it learns (TL). 
CNN-FE learns around 91\% tags of the tagset compared to 73\% with the feature based system.
It is an interesting improvement, because model is learning more tags and it is better at assigning relevant tags to movies.
We observe similar pattern for the rest of the two sets of tags where we select top three and ten tags.
For all the sets, CNN-FE model learns the highest number of tags compared to the other models.
In terms of micro-F1 and tag recall, it does not achieve the highest numbers but performs very closely.\\
\noindent{\textbf{Incompleteness in Tag Spaces}}: One of the limitations of folksonomies is the incompleteness in tag spaces. The fact that users have not tagged an item with a specific label does not imply that that label does not apply to the item. 
Incompleteness makes learning challenging for computational models as the training and evaluation process penalizes the model for predicting a tag that is  not present in the ground truth tags, even though in some cases it may be a suitable tag.
For example, ground truth tags for the movie \textit{Luther (2003)}\footnote{\url{http://www.imdb.com/title/tt0309820}} are \textit{murder, romantic,} and \textit{violence} (Table \ref{tab_example_tag_fp}). 
And the predicted tags from our proposed model are 
\textit{murder, melodrama, intrigue, historical fiction, } and \textit{christian film}. The film is indeed a Christian film\footnote{\url{https://www.christianfilmdatabase.com/review/luther-2}} portraying the biography of Martin Luther, who led the Christian reformation during the 16th century. 
According to the Wikipedia, ``\textit{Luther is a 2003 American-German epic historical drama film loosely based on the life of Martin Luther''}\footnote{\url{https://en.wikipedia.org/wiki/Luther_(2003_film)}}.
Similarly, Edtv\footnote{\url{http://www.imdb.com/title/tt0131369/}} (Table \ref{tab:tab_tags_script_plot_example}) has tags \textit{romantic} and \textit{satire} in the dataset. Our system predicted \textit{adult comedy} and this tag is appropriate for this movie.\\
In these two cases, the system will get lower micro-F1 since the relevant tags are not part of the ground truth. Perhaps a different evaluation scheme could be better suited for this task. We plan to work on this issue in our future work.\\
\noindent{\textbf{Significance of the Flow of Emotions}: } The results suggest that incorporating the flow of emotions helps to achieve better results by learning more tags. 
\begin{figure}[h]
\centering
\includegraphics[width=0.45\textwidth]{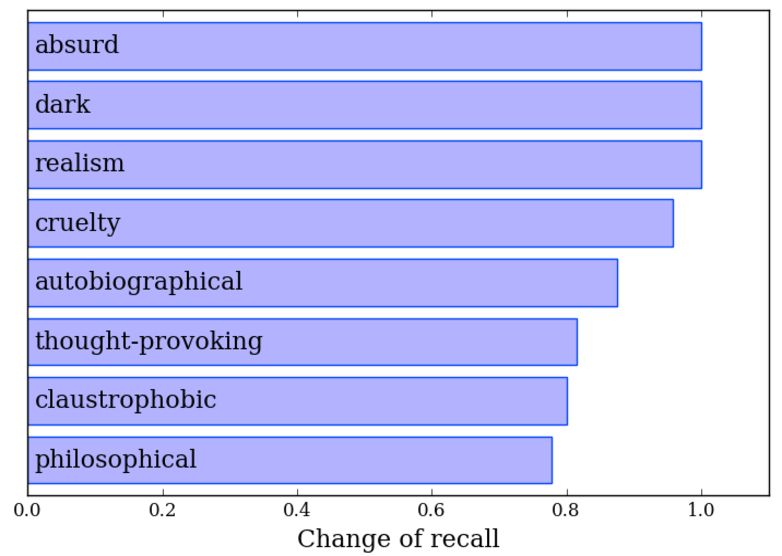}
\caption{\small Tags with higher change of recall after adding the flow of emotions in CNN.}
\label{fig_recall_change}
\end{figure}
\textcolor{black}{
Figure \ref{fig_recall_change} shows some tags with significant improvements in recall after incorporating the flow of emotions.
We notice such improvements for around 30 tags.
We argue that for these tags (e.g. \textit{absurd, cruelty, thought-provoking, claustrophobic}) the changes in specific sentiments are adding new information helpful for identifying relevant tags. 
But we also notice negative changes in recall for around 10 tags, which are mostly related to the theme of the story (e.g. \textit{blaxploitation, alternate history, historical fiction, sci-fi}).
It will be an interesting direction of future work to add a mechanism that can also learn to discern when emotion flow should contribute more to the prediction task.
}
\begin{figure}[h]
\centering
\includegraphics[scale=0.5]{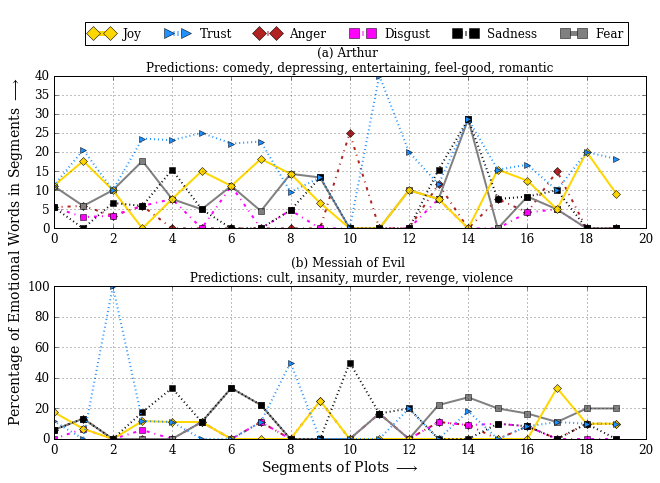}
\caption{\small The flow of emotions in the plots of 2 different types of movies. Each synopsis was divided into 20 segments based on the words, and percentage of the emotions for each segment was calculated using the NRC emotion lexicons. The y axis represents the percentage of emotions in each segment; whereas, the x axis represents the segments.}
\label{fig:emotion_flow}
\end{figure}
~\\
In Figure \ref{fig:emotion_flow}, we inspect how the flow of emotions looks like in different types of plots. 
Emotions like \textit{joy} and \textit{trust} are continuously  dominant over \textit{disgust} and \textit{anger} in the plot of \textit{Arthur (1981)}, which is a comedy film. 
We can observe sudden spikes in \textit{sadness} and \textit{fear} at segment 14, which is the possible reason for triggering the tag \textit{depressing}. 
We observe a different pattern in the flow of emotions in \textit{Messiah of Evil (1973)}, which is a horror film. 
Here the dominant emotions are \textit{sadness} and \textit{fear}. 
Such characteristics of emotions are helpful to determine the type and possible experiences from a movie.
Our model seems to be able to leverage this information that is allowing it to learn more tags; specifically tags that are related to feelings.\\
\noindent{\textbf{Learning or Copying?}}
We found that only 11.8\% of the 14,830 predicted tags for the $\sim$3K movies in the test data were found in the synopses themselves. 12.7\% of the total 9,022 ground truth tags appear in the plot synopses. These numbers suggest that the model is not dependent on the occurrences of the tags in the synopses to make predictions, rather it seems it is trying to understand the plots and assign tags based on that.
We also found that all the tags that were present in the synopses of the test data are also present in the synopses of the training data. 
Then we investigate what type of tags appear in the synopses and which ones do not.
Tags present in the synopses are mostly genre or event related tags like \textit{horror, violence, historical}.
On the other hand, most of the tags that do not appear in the synopses are the tags that require a more sophisticated analysis of the plots synopses (e.g. \textit{thought-provoking, feel-good, suspenseful}).
It is not necessarily bad to predict tags that are in the synopses, since they are still useful for recommender systems. However, if this was the only ability of the proposed models, their value would be limited. 
Luckily this analysis, and the results presented earlier show that the model is able to infer relevant tags, even if they have not been observed in the synopses. This is a much more interesting finding.\\
\noindent{\textbf{Learning Stories from Different Representations: }} Movie scripts represent the detailed story of a movie, whereas the plot synopses are summaries of the movie. \textcolor{black}{The problem with movie scripts is that they are not as readily available as plot synopses. However, it is still interesting to evaluate our approach to predict tags from movie scripts.
\begin{table}[h]
\centering
\small
\begin{tabular}{@{}|l|rrr|rrr|@{}}
 \hline
 & \multicolumn{3}{c|}{\textbf{Top 3}} & \multicolumn{3}{c|}{\textbf{Top 5}} \\  \hline
 & \textbf{F1} & \textbf{TR} & \textbf{TL} & \textbf{F1} & \textbf{TR} & \textbf{TL} \\  \hline
Plot Synopses & 29.3 & 8.04 & 28 & 38.7 & 15.70 & 35 \\
Scripts& 29.8 & 5.16 & 19 & 37.0 & 9.27 & 26 \\ \hline
\end{tabular}
\caption{\small Evaluation of predictions using plot synopses and scripts}
\label{tab_evaluation_plot_scripts}
\end{table} 
For this purpose, we collected movie scripts from our test set. We were able to find 80 movie scripts using the ScriptBase  corpus \cite{scriptbase_gorinsky_2015}.}\\
%
%
In table \ref{tab_evaluation_plot_scripts}, we show the evaluation of tags generated using plot synopses and scripts. Despite having similar micro-f1 scores, \textit{tag recall} and \textit{tags learned} are lower when we use the scripts. \textcolor{black}{A possible explanation for this is the train/test mismatch since the model was trained using summarized versions of the movie, while the test data contained full movies scripts. Additional sources of error could come from the external info included in scripts (such as descriptions of actions from the characters or settings).}
%
\begin{table}[!hbtp]
\centering
\small
\begin{tabular}{|c|c|}
\hline
\textbf{Percentage of Match} & \textbf{Percentage of Movies} \\ \hline
$>=$80\% & 40\% \\ 
$>=$40\% \& $<$80\% & 47.5\% \\
$>=$20\% \& $<$40\% & 11.25\% \\
\hline
\end{tabular}
\caption{\small Percentage of the match between the sets of top five tags generated from the scripts and plot synopses.}\label{tab:table_script_plot_tag_match}
\end{table}\\
Table \ref{tab:table_script_plot_tag_match} shows that for most of the movies we generate very similar tags using the scripts and plot synopses. For 40\% movies, at least 80\% tags are the same. \textcolor{black}{While the predictions are not identical, these results show a consistency in the learned tags from our system. An interesting direction for future work would be to study what aspects in a full movie script are relevant to predict tags.}\\
\begin{table}[!hbtp]
\small





\centering
\begin{tabular}{|l|}
\hline

\begin{tabular}[c]{@{}l@{}}
\textbf{Title: }A Nightmare on Elm Street 5: The Dream Child\\
\textbf{Ground Truths:} cult, good versus evil, insanity, murder, sadist, violence\\
\textbf{Synopsis:} cult, murder, paranormal, revenge, violence\\
\textbf{Script:} murder, violence, flashback, cult, suspenseful\end{tabular} \\ \hline
                                                                                                                                                                                                                                                          \begin{tabular}[c]{@{}l@{}}
\textbf{Title: }EDtv\\
\textbf{Ground Truths:} romantic, satire\\
\textbf{Synopsis:} adult comedy, comedy, entertaining, prank, satire\\
\textbf{Script:} comedy, satire, prank, entertaining, adult comedy \end{tabular} \\ \hline

\begin{tabular}[c]{@{}l@{}}
\textbf{Title: }Toy Story\\
\textbf{Ground Truths:} clever, comedy, cult, cute, entertaining, fantasy, humor, violence\\
\textbf{Synopsis:} comedy, cult, entertaining, humor, psychedelic\\
\textbf{Script:} psychedelic, comedy, entertaining, cult, absurd \end{tabular} \\ \hline

\begin{tabular}[c]{@{}l@{}}
\textbf{Title: }Margot at the Wedding\\
\textbf{Ground Truths:} romantic, storytelling, violence\\
\textbf{Synopsis:} depressing, dramatic, melodrama, queer, romantic\\
\textbf{Script:} psychological, murder, mystery, flashback, insanity \end{tabular} \\ \hline
                              
\end{tabular}

\caption{\small Example of ground truth tags of movies from the test set and the generated tags for them using plot synopses and scripts.}\label{tab:tab_tags_script_plot_example}
\end{table}\\
\noindent{\textbf{Challenging Tags:}} We found that these seven tags: \textit{stupid, grindhouse film, blaxploitation, magical realism, brainwashing, plot twist,} and \textit{allegory}, were not assigned to any movies in the test set. One reason might be that these are very infrequent (around 0.06\% of movies have them assigned as their tags). This will obviously make them difficult to learn. Again, these are  subjective as well. We believe that tagging a plot as stupid or brainwashing is  complicated and depends on perspectives of a tagger. We plan to investigate such type of tags in the future.


%
%
%

 \section{Conclusions and Future Work}\label{sec:conclusion}
In this paper we explore the problem of automatically creating tags for movies using plot synopses. We propose a model that learns word level feature representations from the synopses using CNNs and models sentiment flow throughout the plots using a bidirectional LSTM. We evaluated our model on a corpus that contains plot synopses and tags of 14K movies. We compared our model against a majority and random baselines, and a system that uses traditional hand-crafted linguistic features. We found that incorporating emotion flows boosts prediction performance by improving the learning of tags related to feelings as well as increasing the overall number of tags learned.\\
Predicting tags for movies is an interesting and complicated problem at the same time. To further improve our results, we plan to investigate more sophisticated architectures and explore ways to tackle the problem of incompleteness in the tag space. We also plan to evaluate the quality of predicted tags using a human study evaluation and experiment on predicting tags in other storytelling related domains.

\section*{Acknowledgements}
This work was partially supported by the National Science Foundation under grant number 1462141 and by the U.S. Department of Defense under grant W911NF-16-1-0422.

\bibliographystyle{acl}
\bibliography{coling2018}
\end{document}